\documentclass{article}

\usepackage{microtype}
\usepackage{graphicx}
\usepackage{subcaption}
\usepackage{booktabs} 

\newcommand{\ldotsTwo}{%
  \mathinner{{\ldotp}{\ldotp}}%
}

\usepackage{hyperref}

\usepackage{pgfplots}
\usepgfplotslibrary{groupplots}
\usepackage{tikz}
\usetikzlibrary{matrix}

\definecolor{spectral1}{HTML}{9E0142}
\definecolor{spectral2}{HTML}{D53E4F}
\definecolor{spectral3}{HTML}{F46D43}
\definecolor{spectral4}{HTML}{FDAE61}
\definecolor{spectral5}{HTML}{606669} 
\definecolor{spectral6}{HTML}{FFFFBF}
\definecolor{spectral7}{HTML}{E6F598}
\definecolor{spectral8}{HTML}{ABDDA4}
\definecolor{spectral9}{HTML}{66C2A5}
\definecolor{spectral10}{HTML}{3288BD}
\definecolor{spectral11}{HTML}{5E4FA2}

\usepackage[preprint]{icml2026}

\usepackage{amsmath}
\usepackage{amssymb}
\usepackage{mathtools}
\usepackage{amsthm}

\usepackage[nolist,nohyperlinks]{acronym}
\begin{acronym}
\acro{MHSA}{multi-head self-attention}
\acro{LLM}{large language model}
\acro{NLP}{natural language processing}
\acro{RoPE}{rotary positional embeddings}
\acro{CNN}{convolutional neural network}
\acro{ViT}{Vision Transformer}
\end{acronym}

\usepackage[capitalize,noabbrev]{cleveref}

\theoremstyle{plain}

\theoremstyle{definition}

\theoremstyle{remark}

\usepackage[textsize=tiny]{todonotes}

\icmltitlerunning{Reg4Pru: Token Pruning Regularisation}
\newcommand{\paperacronym}{Reg4Pru}

\begin{document}

\twocolumn[
  \icmltitle{Reg4Pru: Regularisation Through \\ Random Token Routing for Token Pruning}
  





  \icmlsetsymbol{equal}{*}

  \begin{icmlauthorlist}
    \icmlauthor{Julian Wyatt}{CS_Ox}
    \icmlauthor{Ronald Clark}{CS_Ox}
    \icmlauthor{Irina Voiculescu}{CS_Ox}
  \end{icmlauthorlist}

  \icmlaffiliation{CS_Ox}{Department of Computer Science, University of Oxford}

  \icmlcorrespondingauthor{Julian Wyatt}{Julian.Wyatt@cs.ox.ac.uk}

  \icmlkeywords{Token-Routing, Token-Pruning, Regularisation, Segmentation, Attention}

  \vskip 0.3in
]



\printAffiliationsAndNotice{}  

\begin{abstract}

    Transformers are widely adopted in modern vision models due to their strong ability to scale with dataset size and generalisability. However, this comes with a major drawback: computation scales quadratically to the total number of tokens. Numerous methods have been proposed to mitigate this. For example, we consider token pruning with reactivating tokens from preserved representations, but the increased computational efficiency of this method results in decreased stability from the preserved representations, leading to poorer dense prediction performance at deeper layers. In this work, we introduce {\paperacronym}, a training regularisation technique that mitigates token-pruning performance loss for segmentation. We compare our models on the FIVES blood vessel segmentation dataset 
    and find that {\paperacronym} improves average precision by an absolute 46\% compared to the same model trained without routing. This increase is observed using a configuration that achieves a 29\% relative speedup in wall-clock time compared to the non-pruned baseline. These findings indicate that {\paperacronym} is a valuable regulariser for token reduction strategies.

\end{abstract}

\section{Introduction}

\begin{figure}
  \centering

\begin{tikzpicture}
    
            \node[inner sep=0] (image1) at (0,0) {\includegraphics[width=\columnwidth]{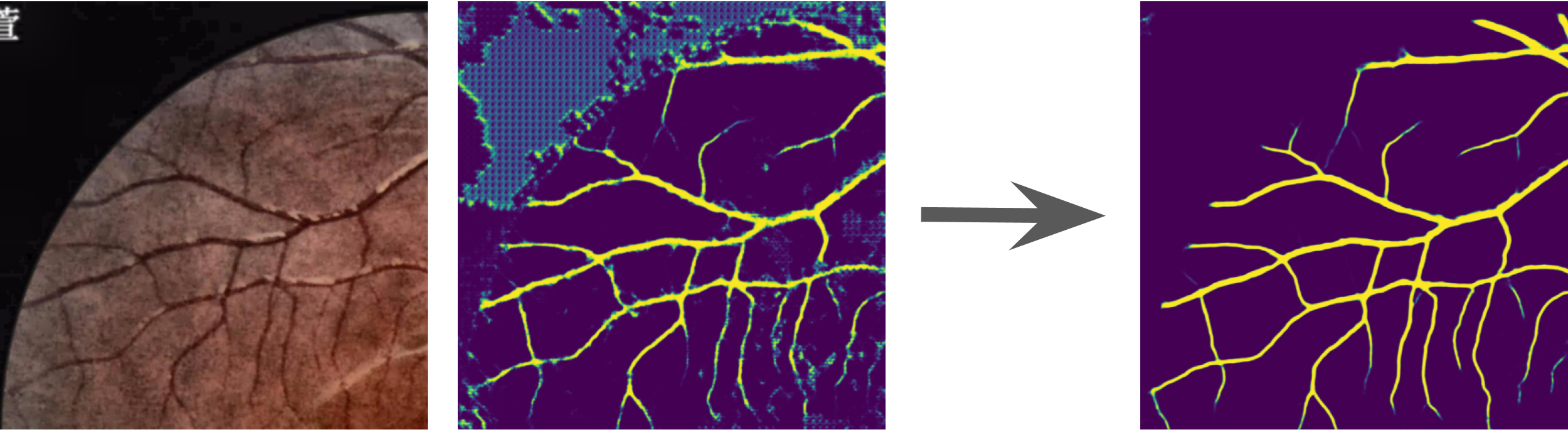}};

            \node[text width=2cm] at (-2.7,1.3)  {\footnotesize	 Input image};
            \node[text width=2cm] at (-0.4,1.3)  {\footnotesize	 No Routing};
            \node[text width=1.5cm] at (3.25,1.3)  {\footnotesize	 {\paperacronym}};

        \end{tikzpicture}

    \begin{tikzpicture}
    
    \begin{axis}[
        width=0.75\columnwidth,height=4cm,
        xlabel={Throughput (imgs/sec)},
        ylabel={\small Average Precision},
        xtick pos=left,
        ytick pos=left,
        extra x ticks={0,1000},
        extra y ticks={0,1},
        xmin = 75,
        xmax = 116, 
        ymin = 0,
        ymax = 0.8, 
        xtick distance=10,
        solid,
        enlarge x limits=false,
        every x tick/.style={color=black, thin},
        every y tick/.style={color=black, thin},
        y tick label style={font=\small},
        x tick label style={font=\small},
        tick align=outside,
        ylabel near ticks,
        axis on top,
        every axis plot/.append style={
            very thick,
            mark size=1.2pt
        },
        xmajorgrids,
        ymajorgrids,
        extra y ticks={0},
        extra y tick style={
          grid=none,
        },
        legend columns=1,
          legend style={
            at={(1.02,0.5)},      
            anchor=west,
            draw=none,
            font=\footnotesize,
            row sep=2pt,
            column sep=0.75em
          },
          legend cell align=left,
    ]

    \addplot [
        spectral4,
        solid,
        very thick,
        mark=*,
        mark size=0.9pt,
        mark options={solid}
    ] table [x=THROUGHPUT, y=AP, col sep=comma]
    {Graph_Data/Pruned_Mask_vs_Throughput.csv};
    \addlegendentry{\textbf{(a)}}\label{plots:RandomRouting}

    \addplot [
        spectral2,
        solid,
        very thick,
        mark=*,
        mark size=0.9pt,
        mark options={solid}
    ] table [x=THROUGHPUT, y=AP, col sep=comma]
    {Graph_Data/Pruned_Mask_Fixed_Route_vs_Throughput.csv};
    \addlegendentry{\textbf{(b)}}\label{plots:FixedRouting}

    \addplot [
        spectral11,
        solid,
        very thick,
        mark=*,
        mark size=0.9pt,
        mark options={solid}
    ] table [x=THROUGHPUT, y=AP, col sep=comma]
    {Graph_Data/No_Routing_vs_Throughput.csv};
    \addlegendentry{\textbf{(c)}}\label{plots:NoRouting}


    \addplot [
        spectral10,
        solid,
        very thick,
        mark=*,
        mark size=0.9pt,
        mark options={solid}
    ] table [x=THROUGHPUT, y=AP, col sep=comma]
    {Graph_Data/ToMe_MLP_Attn_vs_Throughput_reduced.csv};
    \addlegendentry{\textbf{(d)}}\label{plots:ToMe_MLP}

    \addplot[
      gray!90,
      dashed,
      very thick,
      mark=x,
      mark size = 2.5pt,
      mark options={
        solid,
        black               
      },
      unbounded coords=jump
    ] coordinates {
      (70,0.74)
      (78.4,0.74)   
      (150,0.74)
    };
    \addlegendentry{\textbf{(e)}}\label{plots:noPrune}


    \end{axis}
\end{tikzpicture}
\caption{
\textbf{Top:} Output heatmaps for pruned models with and without routing. 
\textbf{Bottom:} Average precision (AP) vs.\ throughput (H100, batch size 1, $1024^2$ input). 
Key: \textbf{(a)} {\paperacronym}, \textbf{(b)} Pruned (Fixed Routing), \textbf{(c)} Pruned (No Routing), \textbf{(d)} No Pruning, \textbf{(e)} ToMe \cite{bolya2023ToMeDiff} over {\tt MHSA+MLP}. 
Pruning keep ratios range from $0.9$ to $0.5$; ToMe merge ratios range from $0.7$ to $0.4$. All metrics are computed from the final block.
}
\label{fig:AP/TH_TLDR}
\end{figure}
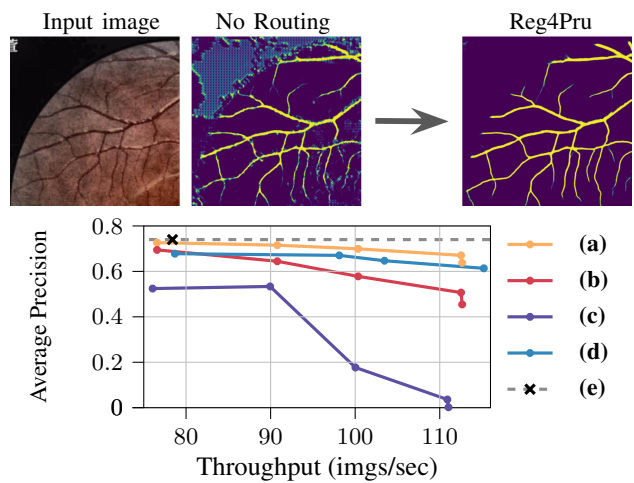

The Transformer architecture~\cite{vaswani2017attention} has become a key foundational block for building modern deep learning models due to their impressive performance when scaling dataset size, and its ability to generalise across input domains. However, the high-quality representations are largely attributed to the attention mechanism which is bottlenecked with $\mathcal{O}(N^2)$ computations. This causes computation to  scale quadratically when increasing tokenised data, over $N$ total tokens. Medical imaging segmentation requires high-resolution images to preserve detail, as lower resolution compromises invaluable details in the image that a doctor would need. Therefore, also motivated to maximise throughput to expedite the analysis for a doctor.

\begin{figure*}[t]

      \centering
      \begin{tikzpicture}
    
            \node[inner sep=0] (image1) at (0,0) {\includegraphics[width=0.95\linewidth]{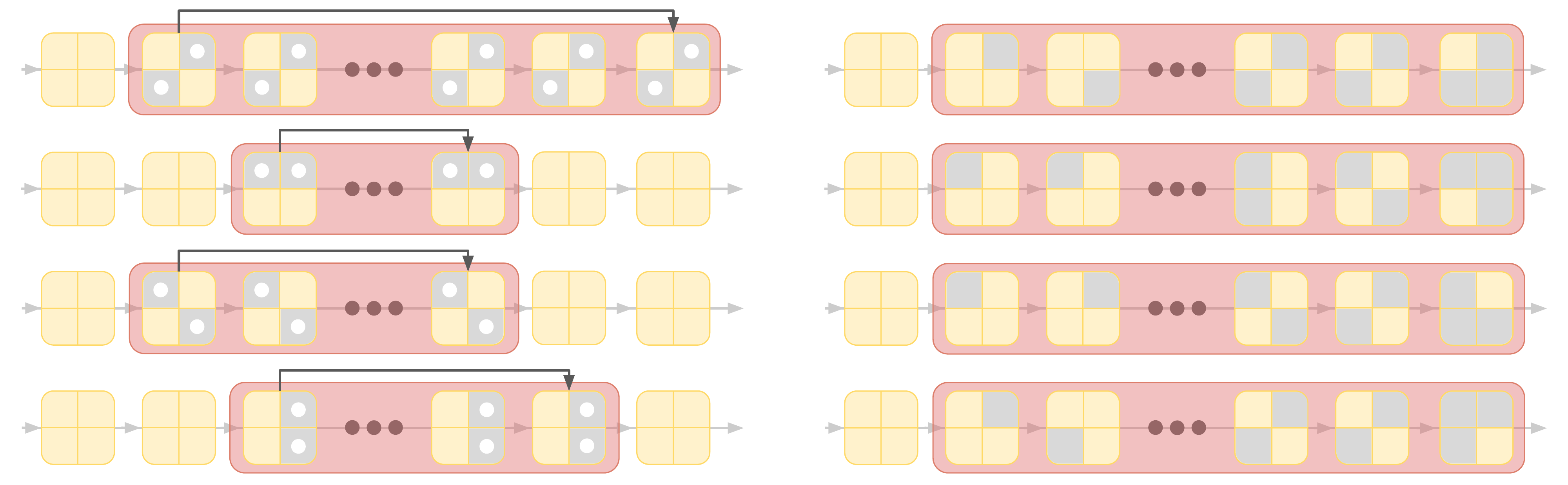}};
            \node[inner sep=0] (image1) at (0,-4) {\includegraphics[width=0.85\linewidth]{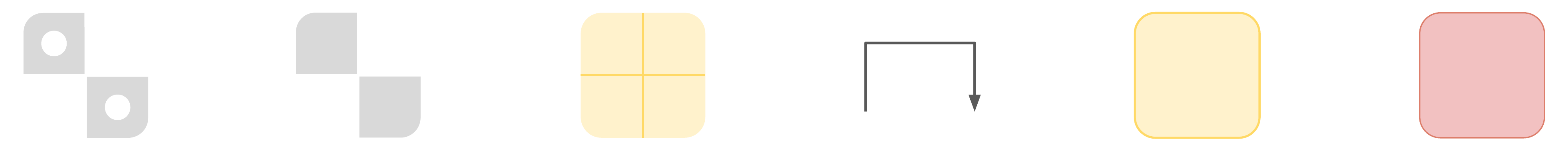}};
    

             \node[rotate=90, text width=1.3cm] at (-8.2,2.2)  {\scriptsize	 Iter 1};
             \node[rotate=90, text width=1.3cm] at (-8.2,0.9)  {\scriptsize	 Iter 2};
             \node[rotate=90, text width=1.3cm] at (-8.2,-0.35)  {\scriptsize	 Iter 3};
             \node[rotate=90, text width=1.3cm] at (-8.2,-1.6)  {\scriptsize	 Iter 4};

            \node[minimum height=0.5cm,text width=2cm] at (-3.9,-3.05)  {\textbf{(a)} Training};
            \node[minimum height=0.5cm,text width=2cm] at (4.5,-3.05)  {\textbf{(b)} Inference};


            \node[minimum height=0.5cm,text width=2cm] at (-7.35,-2.6)  {\tiny Block};
            \node[minimum height=0.5cm,text width=2cm] at (-6.35,-2.6)  {\tiny 0};
            \node[minimum height=0.5cm,text width=2cm] at (-5.35,-2.6)  {\tiny 1};
            \node[minimum height=0.5cm,text width=2cm] at (-4. 3,-2.6)  {\tiny 2};

            \node[minimum height=0.5cm,text width=2cm] at (-2. 57,-2.6)  {\tiny  $L{-}2$};
            \node[minimum height=0.5cm,text width=2cm] at (-1.5,-2.6)  {\tiny  $L{-}1$};
            \node[minimum height=0.5cm,text width=2cm] at (-0.25,-2.6)  {\tiny  $L$};


            \node[minimum height=0.5cm,text width=2cm] at (1.95,-2.6)  {\tiny 0};
            \node[minimum height=0.5cm,text width=2cm] at (3,-2.6)  {\tiny 1};
            \node[minimum height=0.5cm,text width=2cm] at (4.05,-2.6)  {\tiny 2};

            \node[minimum height=0.5cm,text width=2cm] at (5.78,-2.6)  {\tiny  $L{-}2$};
            \node[minimum height=0.5cm,text width=2cm] at (6.87,-2.6)  {\tiny  $L{-}1$};
            \node[minimum height=0.5cm,text width=2cm] at (8.1,-2.6)  {\tiny  $L$};

            \node[minimum height=0.5cm,text width=2cm, align=center] at (-6.45,-5.05)  {\small Routed tokens $\tau_r$};
            \node[minimum height=0.5cm,text width=2cm, align=center] at (-3.9,-5.05)  {\small Pruning policy $\mathbf{P}$};
            \node[minimum height=0.5cm,text width=2cm, align=center] at (-1.3,-5.05)  {\small Total \\tokens $N$};
            \node[minimum height=0.5cm,text width=2cm, align=center] at (1.325,-5.05)  {\small Route \\ $r_{l\rightarrow n}$};

            \node[minimum height=0.5cm,text width=2cm, align=center] at (3.83,-5.05)  {\small ViT\\ Block $l$};

            \node[minimum height=0.5cm,text width=2.5cm, align=center] at (6.5,-5.05)  {\small Blocks not using $\tau_{\{r|p\}} \subset N$};


        \end{tikzpicture}

        
      \caption{\textbf{{\paperacronym} architecture.} \textbf{(a)} - A random 50\% of total tokens $N$ are selected as $\tau_r$ to be routed between uniformly sampled intermediate blocks $[l,n]$, giving the route $r_{l\rightarrow n}$.  \textbf{(b)} - During inference, a policy network makes binary predictions $\mathbf{P}$ for tokens to prune $\tau_p$ that increase hierarchically with the pruning ratio. Following the final block $L$, the tokens are convolutionally decoded with query tokens to give full resolution mask logits.}
      \label{fig:Architecture}
    \end{figure*}
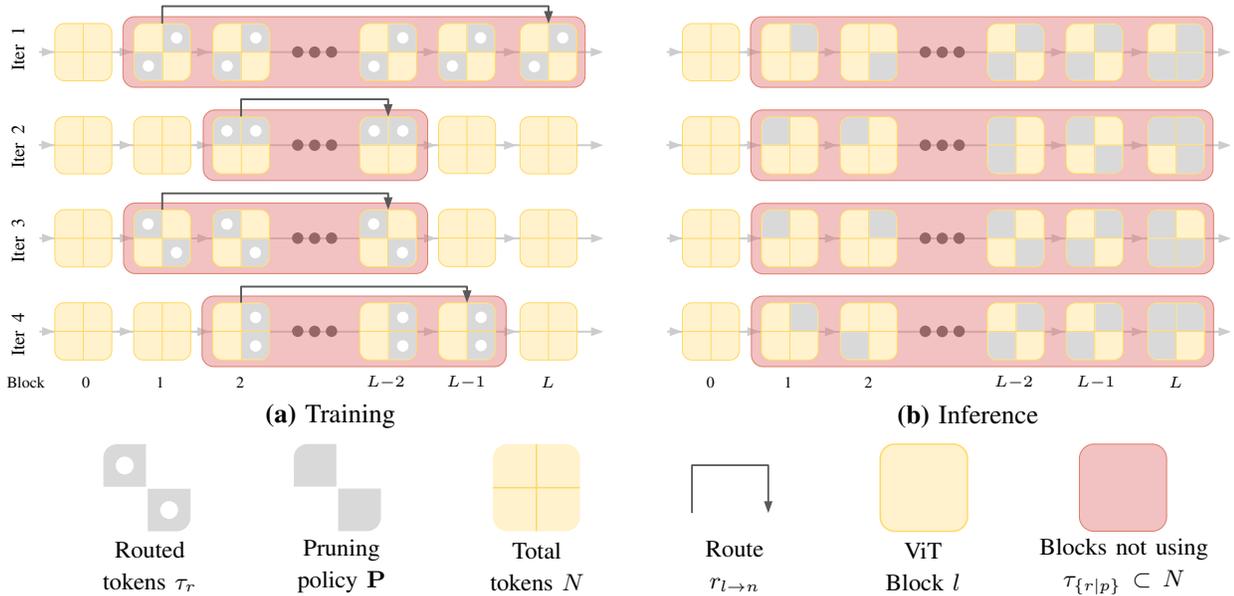

The main imaging variant: the \acf{ViT} \cite{ViT}, tokenises the image by splitting the image into patches and then flattening this representation. Images typically have large areas that are semantically similar, and, therefore, are redundant to consider for computation. As such, the most direct methods to improve efficiency and throughput reduce the quantity of tokens while leaving the original backbone the same.

Pruning, Pooling and Merging make up the core efficiency approaches to cut computation through reducing total token counts throughout the Transformer. \emph{Pruning}~\cite{rao2021dynamicvit}, removes redundant and less useful tokens. \emph{Pooling}~\cite{marin2023token}, utilises K-Means clustering to combine similar tokens. \emph{Merging}~\cite{bolya2023ToMe, bolya2023ToMeDiff}, generates a bipartite graph of source and destination tokens based on cosine similarity that are merged for computation of a module and unmerged afterwards.




In this work, we focus on reducing the reliance on tokens for high resolution dense prediction through pruning. As each token is required for pixel-wise predictions such as segmentation, at inference we reactivate pruned tokens \cite{SVIT} to generate the dense prediction. We specifically study pruning-based efficiency methods that temporarily remove tokens and later reintroduce them during inference. However, upon evaluating this, we observe a degradation of performance at deeper layers. This is counter-intuitive, since model performance typically increases with model depth. The instability appears as a train-test distribution difference, leading pruned tokens to have drastically different representations to what they would have at that block during training. This is explicitly shown at the top of \cref{fig:AP/TH_TLDR}. In order to mitigate this phenomenon, we mimic the effect of pruning, during training, by applying token routing \cite{raposo2024mixture,krause2025tread}. This concept selects tokens to skip the following blocks until the end of the route. While this begins to regularise the model, it still leaves instability outside of the routing range; see \cref{fig:depthwise}. Therefore, by randomising the routing range, we refine the effect of pruning during training, and consequently improve the overall representation at the final layer. The general train-time routing and inference-time pruning structure can be seen in \cref{fig:Architecture}.



\newpage

\textbf{Contributions} - In summary, we make the following contributions:
\begin{itemize}
    \item  We perform a depth-wise analysis on the token pruning performance for dense prediction and observe an instability for pruned tokens.
    \item To mitigate this instability, we propose train-time token routing with random bounds, {\paperacronym}. 
    \item We find that this proposition improves the stability for frequently pruned tokens, improving the average precision on the FIVES dataset. 
\end{itemize} 

\section{Related Work}

\textbf{\Acfp{ViT}} - Largely inspired by several works from the \ac{LLM} community \cite{vaswani2017attention, devlin2019bert, brown2020language}, the Transformer was rapidly translated to vision applications \cite{ViT}. As a consequence to the emphasis on efficiency and performance improvements in \acp{LLM}, mechanisms continue to be inherited from \ac{NLP} research. For example, \ac{RoPE} \cite{su2024roformer, heo2024rotary}, Key-Value Caching \cite{shi2025scalingPS3}, and Token Routing \cite{raposo2024mixture,krause2025tread} have recently been adapted for \acp{ViT}. 



\textbf{Dense Predictions} - Tokens that capture long-range context across the image naturally excel at dense prediction tasks. However, despite their strong representations, many works incorporate convolutional encoders or decoders \cite{carion2020endDETR, vitadapter, mask2former, ranftl2021visionDPT, SEGFORMER} to enhance local context and hierarchical feature extraction. This often results in different output blocks being better suited for dense prediction than for classification \cite{bolya2025perception}. More recently, \cite{kerssies2025your} propose simplifying transformers for dense prediction by removing adapters and hierarchical components, instead generating intermediate mask predictions to aid optimisation. In contrast, by increasing the similarity between block outputs, our work enables efficient dense prediction using only the final block’s output. Overall, these approaches demonstrate that while Transformers are well suited to dense prediction, achieving efficiency in this setting introduces additional challenges.

\textbf{Efficient Transformers} - Transformer efficiency can be improved through a range of strategies, including neural architecture search \cite{JMLR:v20:18-598NeuralArchitectureSearch, liu2022neural}, pruning of layers or attention heads \cite{shim2021layer, he2024matters}, and quantisation \cite{ma2024era}. Other works focus on reducing the computational complexity of attention from $\mathcal{O}(N^2)$ to $\mathcal{O}(N\log N)$ \cite{kitaev2020reformer} or $\mathcal{O}(N)$ \cite{katharopoulos2020transformers}. In this work, we focus on token reduction strategies such as pruning and merging, which directly reduce computation while largely preserving the underlying model structure.

The seminal work on \ac{ViT} token pruning, DynamicViT \cite{rao2021dynamicvit}, hierarchically reduces the total token counts to reduce computation. They use a lightweight prediction module to predict the most informative tokens, and drop all remaining tokens. Token Merging \cite{bolya2023ToMe} selects two subsets from the total tokens, then reduces the two subsets by averaging tokens from the first set with their most similar token in the second set.  

Unlike for classification, token reduction strategies for dense predictions are less trivial as each token is required for pixel-wise predictions at inference. Token Merging for Diffusion \cite{rombach2022high, bolya2023ToMeDiff} simplifies the merging strategy by unmerging tokens after each computation block. \citet{tang2023dynamic} propose early exiting of highly confident tokens with multiple dense predictions that are subsequently combined at the end. \citet{SVIT} question hierarchical pruning for dense predictions and instead propose reactivating pruned tokens. Most pruning methods optimise per-token predictions independently; however, \citet{zhou2023token} propose a soft top-$k$ formulation based on Gumbel-Max to improve full-image predictions.

\textbf{Token Routing} - Beyond token reduction, recent work has explored propagating tokens between layers to improve training and inference efficiency. Originally \citet{raposo2024mixture} proposed using a learnt routing mechanism that decides whether a subset of tokens will be routed. The routing mechanism is then applied at both training and inference to gain efficiency improvements. TREAD \cite{krause2025tread} propose using random token routing during training to cut GPU train hours, enabling considerably quicker training of diffusion models. Our work builds on these concepts by extending token routing to regularise dense predictions with token pruning. 

\section{Method}
\subsection{Preliminary} \label{sec:method-prelim}

\textbf{\Acfp{ViT}} - \acp{ViT} \cite{ViT} take an input image $\mathit{I} \in \mathbb{R}^{3\times H \times W}$ and split the image into $N$ non-overlapping patches of shape ($H_p$, $W_p$) where $H$ and $W$ are the image's height and width and ($H_p$, $W_p$) is the height and width of the patches. These patches are then linearly projected and reshaped into tokens $\mathbf{\tau^0} \in \mathbb{R}^{N\times D}$, with dimensionality $D$. Each token is subsequently processed by $L$ transformer blocks \cite{vaswani2017attention}. Each transformer block includes a $QKV$ multi-head self-attention layer ({\tt MHSA}) and a multi-layer perceptron ({\tt MLP}), where $QKV$ are 3 different projections of the input for self-attention. Formally, each block $l$ applies the following:
\begin{equation}
\begin{aligned}
Z^l &= \tau^l + \texttt{MHSA}(\texttt{Norm}(\tau^l)) \\
\tau^{l+1} &= Z^l + \texttt{MLP}(\texttt{Norm}(Z^l))
\end{aligned}
\end{equation}
where {\tt Norm} is Layer Normalisation \cite{ba2016layer}. The 
{\tt MHSA} layer is the core component that boosts expressivity of the Transformer, however, it does so with $\mathcal{O}(N^2)$ operations when calculating $QK^T$. In many applications which require high-resolution imaging: dense prediction or QA tasks \cite{mask2former, shi2025scalingPS3}, scaling up resolution is vital to improve performance. Token pruning directly addresses this to increase throughput.   


\textbf{Pruning} - The majority of pruning methods optimise a hierarchy of binary policies $\mathbf{P}_\tau^s \in \{0,1\}^N$ at stage $s$, for tokens $\tau$, sampled from $p_\tau$, that determines which tokens $\tau_p \subset N$ will be pruned in subsequent blocks. A stage $s$ is the group of consecutive transformer blocks that share the same pruning decision $\mathbf{P}_\tau^s$. To learn this, recent works use a ratio-based loss to enforce a predetermined budget and dynamically infer which tokens are relevant; however, this requires re-training for a new budget. \citet{rao2021dynamicvit} propose using a per-image ratio loss, while \citet{kong2022spvit} proposed a batch-wise ratio loss to allow for an adaptive budget between images. Ratio budgets are applied hierarchically where, for a keep ratio at an initial stage $\rho_1$, the following stages typically use a budget of $\rho^s$ as this reduces FLOPs by approximately $\rho$. Where listed, we use the per-image ratio loss, for batch size $B$, pruning stages $S$, and budget for a given stage $\rho_s$ and by default $\rho=0.7$, \begin{equation}
    \mathcal{L}_{\text{ratio}} = \frac{1}{BS}\sum^B_{b=1}\sum^S_{s=1}\left(\rho_s - \frac{1}{N}\sum^N_{\tau=1}\mathbf{P_\tau^{b,s}}\right)^2
\end{equation} as training on large images (e.g. $>1024^2$) usually requires a small batch size due to memory budgets, reducing the effect of the adaptive budget.

\textbf{Differentiability} - To enforce this objective, we require the sampling from token-wise policy predictions $p_\tau$ to binary policies $\mathbf{P}\tau$ to be differentiable.
However, the use of \textit{argmax} to discretise the policy predictions $p_\tau$ is non-differentiable. Additionally, removing tokens during training consequently cuts the gradients for those tokens at the point of pruning. \citet{jang2017categorical} proposed a differentiable alternative, Gumbel-Softmax, which reparameterises the discrete sampling of $p_\tau$ into stochastic continuous sampling. Then, by applying the Straight-Through Gumbel-Softmax variant, we get a discrete output. As such, we generate per-token binary policy decisions with $$\mathbf{P}_\tau = \text{Gumbel-Softmax}(p_\tau) \in \{0,1\}^N.$$  Moreover, we stick to the independent, per-token predictions \cite{rao2021dynamicvit} as we found it to be more stable than soft top-$k$ Gumbel implementations \cite{zhou2023token} applied over the full token sequence $N$. While this allows an optimisation over the choice of tokens to be pruned during training, this still does not prevent inter-token communication during self attention. Therefore, we use Attention Masking \cite{rao2021dynamicvit} which differentiably mimics pruning during training, while leaving all tokens in the forward pass, and allowing for end-to-end training. This applies the policy mask in exp-space during the attention softmax \cite{rao2021dynamicvit}. Conversely, instead of in logit-space such as with FlexAttention score functions \cite{dong2024flex}. This redefines the attention softmax as: 
\begin{equation}
    \mathbf{M}_{ij} =
    \begin{cases}
        1, & i = j,\\
        \mathbf{P}_j, & i \neq j,
    \end{cases}
    \qquad 1 \le i,j \le N, \; i,j \subset \tau,
    \label{eq:policy_to_attention_mask}
\end{equation}
\begin{equation}
    \mathbf{A}_{ij} = QK^T / \sqrt{d_k},
\end{equation}
\begin{equation}\label{eq:pruned_attention}
    \text{Softmax}(\mathbf{A}) = \frac{\text{exp}(\mathbf{A}_{ij})\mathbf{M}_{ij}}{\sum^N_{k=1}\text{exp}(\mathbf{A}_{ik})\mathbf{M}_{ik}}.
\end{equation}
Here, $\mathbf{M}$ converts the policy mask $\mathbf{P}$ into the symmetric attention mask via \cref{eq:policy_to_attention_mask}, which adds an additional self-loop to improve the numerical stability, by ensuring a token can attend to itself. Further, $A$ is the attention output \cite{vaswani2017attention}.

\textbf{Inference} - 
With the full capacity for end-to-end training, we are now able to reduce the inference-time compute by selecting our policy from a top-$k$ over all token predictions $p_\tau$ and setting the binary policy mask $\mathbf{P_\tau}$ to this decision. This activates $\tau_p \subset N$ over each \ac{ViT} block until the next token predictions $p_\tau$. This is unlike \cite{SVIT}, which instead takes an \textit{argmax} for a dynamic pruned token count.

\textbf{Architecture} - For dense predictions, it is preferable to maintain all tokens in memory and instead gather the tokens chosen by the policy and scatter back the output of the block \cite{SVIT}. Furthermore, this idea lends itself very well to using the EoMT architecture \cite{kerssies2025your}. EoMT optimises transformer dense predictions by removing convolutional adapters \cite{vitadapter, mask2former} and trains with the standard Mask2Former-style output and loss prediction \cite{mask2former}.


\subsection{Train-time Token Routing with
Random Bounds}

Motivated by the depth-wise instability observed under token pruning with reactivation (\cref{fig:AP/TH_TLDR}), we now describe our train-time token routing strategy which promotes depth-invariant features for learning pruning policies. Following the ideas of token routing from Mixture-of-Depths \cite{raposo2024mixture} and TREAD \cite{krause2025tread}, we introduce {\paperacronym}, which acts as a train-time regulariser. Token routing, introduced to improve LLM efficiency, bypasses a subset of tokens for specific blocks, and is a natural candidate to simulate pruning during training. Notably, in our framework, we keep routing unparameterised with randomly sampled bounds at each training forward pass. This induces a generalisation for pruned tokens between training and testing, where the inference-time pruning predictions are governed by a separate token pruning policy module.

\begin{figure}[h]

      \centering
      \begin{tikzpicture}
    
            
            \node[inner sep=0] (image1) at (0,0) {\includegraphics[width=0.4\textwidth]{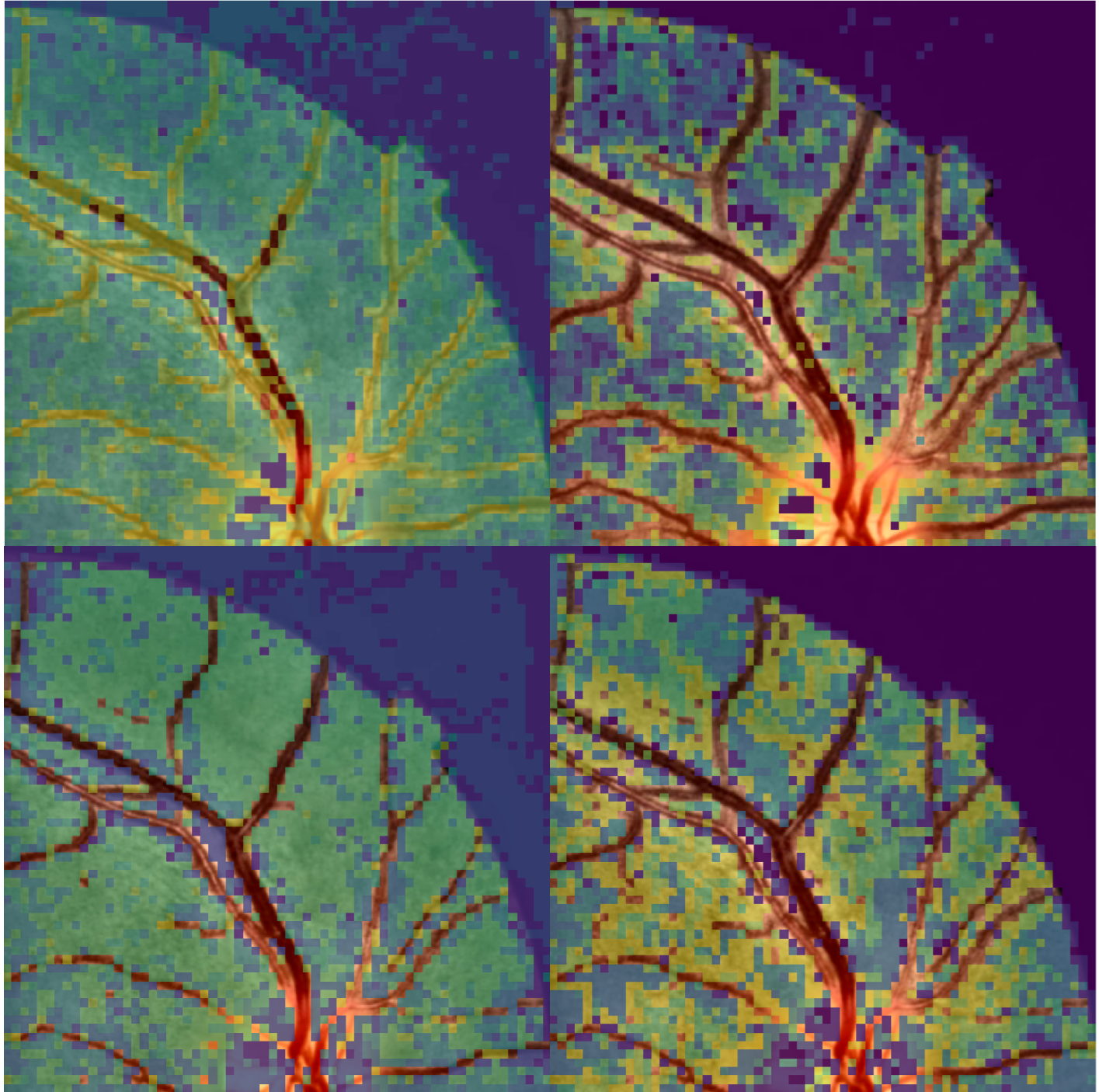}};

    

            \node[rotate=90, minimum height=0.5cm,text width=2cm] at (-3.7,2.4)  { $\mathcal{L}_\text{ratio}$};
            \node[rotate=90, minimum height=0.5cm,text width=2cm] at (-3.7,-1)  {$\mathcal{L}_\text{pol}$};

            \node[minimum height=0.5cm,text width=2cm] at (-1.6,3.7)  {No Routing};
            \node[minimum height=0.5cm,text width=3cm] at (2.5,3.7)  { {\paperacronym}};

        \end{tikzpicture}

      \caption{\textbf{Selected Pruning Policy Frequency} - Frequency with which each token is selected across blocks $[3\ldotsTwo 11]$ (zero-indexed), overlaid on the input image (also shown in \cref{fig:qualitative_logit_routing_ablation} row 1). Tokens shown as fully transparent are selected at every block. Remaining frequencies are visualised using the viridis colourmap, where darker values indicate less frequent selection and lighter values indicate more frequent selection. Examples shown for models trained with the ratio loss $\mathcal{L}_{\text{ratio}}$, and $\mathcal{L}_{\text{pol}}$ where random routing ({\paperacronym}) is used or no routing is used.}
      \label{fig:chosen_policy}
    \end{figure}
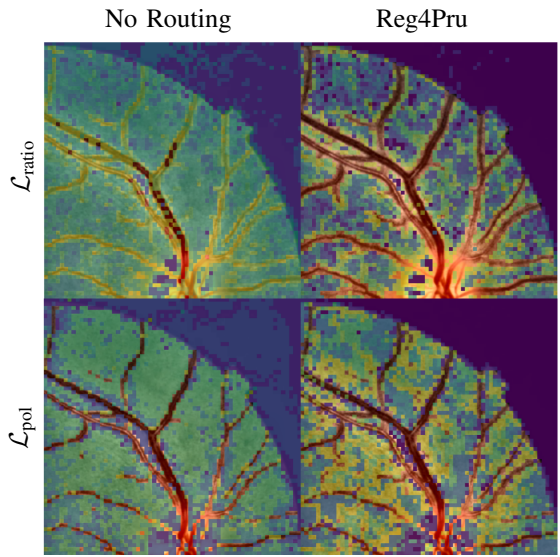

In this work, at the beginning of the forward pass, we select a routing range $[l\ldotsTwo n]$. When computation reaches block $l$, we randomly select a subset of tokens to keep $\tau_k \subset N$, that are gathered from the full set of tokens $N$. The remaining tokens $\tau_r \subset N, \tau_r \cap \tau_k = \varnothing$ skip the following transformer blocks until block $n$, and form the full route $r_{l\rightarrow n}$. The summary of this can be seen in \cref{fig:Architecture}. Therefore, \cref{eq:pruned_attention} and the rest of the block calculations are calculated only on $\tau_k$. Specifically, we scatter the output $\tau_k^n$ back into the stale full token features $\tau^l$ using $\tau_k$'s original indices and then continue to compute $\tau^{n+1}$ at the following block on the full token representation. As a consequence of routing during training, the gradients from routed tokens, with respect to routed blocks are cut. However, by randomising both the routed token subset $\tau_r$ and the routing span $[l\ldotsTwo n]$, different tokens contribute gradients to different blocks across iterations, mitigating this effect.

\begin{figure*}[t]

      \centering
      \begin{tikzpicture}
    
            
            \node[inner sep=0] (image1) at (0,0) {\includegraphics[width=\linewidth]{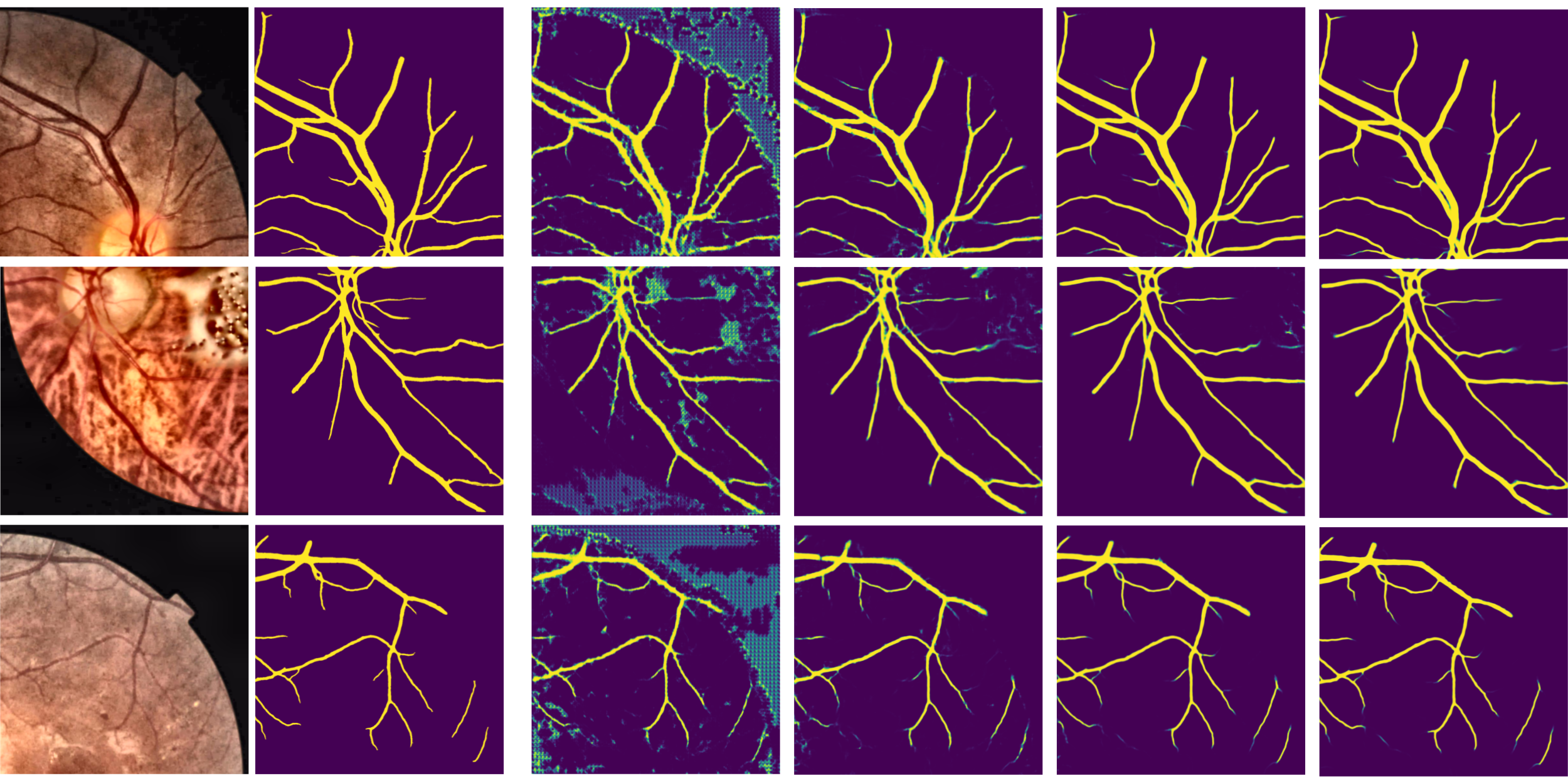}};

    

            \node[minimum height=0.5cm,text width=2cm] at (-6.5,4.45)  { Input};
            \node[minimum height=0.5cm,text width=2cm] at (-3.75,4.45)  {GT};
            \node[minimum height=0.5cm,text width=2cm] at (-1.25,4.45)  {No Routing};

            \node[minimum height=0.5cm,text width=3cm] at (1.95,4.45)  { Fixed Routing};
            \node[minimum height=0.5cm,text width=3cm] at (5.2,4.45)  { {\paperacronym}};
            \node[minimum height=0.5cm,text width=3cm] at (8,4.45)  { No Prune};

        \end{tikzpicture}

      \caption{\textbf{Qualitative segmentation results} - Final block segmentation logits after combining the weighted combination of logits across query masks. Each row contains a distinct input from the FIVES dataset \cite{FIVES}. While each column highlights no routing, fixed bounds routing and randomised routing {\paperacronym} along with a baseline without any pruning or routing.}
      \label{fig:qualitative_logit_routing_ablation}
    \end{figure*}

While computing the remaining tokens $\tau_k$, we are simultaneously learning the pruning policy $\mathbf{P}$. To obtain pruning policy predictions over the full representation of $N$, we scatter $\tau^m, l{\leq}m{\leq}n$ back to generate $p_\tau$. As any token can be reactivated at inference, we want to emulate this and improve the representation for reactivating any pruned token using a misaligned representation of $\tau^l$ at block $n$. Therefore, during training, the token prediction head uses a mixture of current representations $\tau_k$, and stale representations $\tau_r$. Furthermore, if we fix the range $[l\ldotsTwo n]$ for the whole of training \cite{krause2025tread}, we overfit the regularisation to the fixed range; by instead randomising $l$ and $n$, we approximate the feature distribution across pruning depths during training, allowing the tokens to learn depth-wise representation invariance.


\subsection{Informed Policies}

We discussed optimising policies for a specific compute budget  in \cref{sec:method-prelim}; we now consider the idea of learnt informed policies. While potentially being sub-optimal for the primary performance metric, informed policies enable a single predicted policy $\mathbf{P}_\tau$ over any pruning budget. This, allows a single model to support multiple pruning ratios at inference without retraining the model; ideal for deployment to environments that typically require pruning setups, for example, on edge devices \cite{kong2022spvit}.

To learn this policy, we reshape the policy $p_\tau$ back into its grid layout from the patch embedding, select the target as the bilinearly downsampled ground truth segmentation map and optimise with binary cross entropy loss (BCE). Formally, for a target-informed policy $\mathbf{T}$ and pre-Gumbel predictions $p_\tau$, we add \cref{eq:informed_policy} to the total loss.
\begin{equation}\label{eq:informed_policy}
\mathcal{L}_{\text{pol}} = \lambda_{\text{pol}} \frac{1}{B}\sum_{b=1}^{B}\sum_{s=1}^{S} \mathrm{BCE}\big(\sigma(p^{b,s}), \mathbf{T^b}\big)
\end{equation}
The ground truth segmentation map was selected as the default $\mathbf{T}$ as it gives a reasonable prior of foreground vs redundant tokens for pruning setups. $\mathbf{T}$ can be seen in \cref{fig:qualitative_logit_routing_ablation} column 2. While \cref{fig:chosen_policy} highlights how the selection of policy loss informs the underlying selected tokens at inference. 

Immediately before a pruned block, we generate $p_\tau$ from a small prediction network. Prior works employ lightweight token-wise predictors, such as a local-global prediction network \cite{rao2021dynamicvit} or a simple two-layer {\tt MLP} \cite{SVIT}. Crucially, \citet{SVIT} demonstrate that high-capacity predictors are not required for effective and efficient policy selection. Following this study, we adopt a similar network with minor architectural adjustments to improve numerical stability, including the use of RMSNorm \cite{zhang2019root} and layer scaling \cite{touvron2021going}. These changes are not intended to increase the expressivity of the predictor, but instead to ensure stable compatibility with $\mathcal{L}_{\text{pol}}$ while keeping computational costs negligible relative to the backbone. Consequently, we do not ablate and optimise the predictor architecture as it is not a performance bottleneck.

\section{Experiments}

\begin{table*}[t]
  \caption{
    \textbf{Segmentation performance after the final block on FIVES.}
    Comparison of token pruning and token merging (ToMe) methods.
    Best results are shown in \textbf{bold}; $^{*}$ indicates best performance that does not introduce stale representations by reactivating tokens.
    }
  \label{tab:main_results}
  \begin{center}
    \begin{small}
      \begin{sc}
        \begin{tabular}{lcccc}
          \toprule
          Model & \begin{tabular}[l]{@{}l@{}}Base Keep\\ Ratio \end{tabular} & Dice (\%) & \begin{tabular}[c]{@{}l@{}}Average\\ Precision (\%) \end{tabular} & \begin{tabular}[c]{@{}l@{}}Throughput \\ (imgs/sec)\end{tabular} \\ \midrule
        Not Pruned \cite{kerssies2025your}              & -    & 91.01 & 74.31 & 78.49    \\
        ToMe \cite{bolya2023ToMeDiff} - {\tt MHSA}    & 50\% & $90.21^*$ & $71.97^*$ & 102.94   \\
     ToMe \cite{bolya2023ToMeDiff} - {\tt MHSA+MLP} & 50\% & 87.65 & 64.81 & \textbf{103.47} \\
        Pruned $\mathcal{L}_{\text{ratio}}$ No Routing  & 70\% & 70.20 & 25.60 & 100.53   \\ \midrule
        Pruned $\mathcal{L}_{\text{pol}}$ No Routing    & 70\% & 68.68 & 17.70 & 100.95   \\
        Pruned $\mathcal{L}_{\text{pol}}$ Fixed Routing & 70\% & 85.77 & 57.76 & 100.86 \\ \midrule
        Pruned $\mathcal{L}_{\text{ratio}}$ with {\paperacronym}       & 70\% & \textbf{90.46} & \textbf{71.67} & 100.37   \\
        Pruned $\mathcal{L}_{\text{pol}}$ with {\paperacronym}          & 70\% & 89.79 & 69.94 & 100.35  \\ \bottomrule
        \end{tabular}
      \end{sc}
    \end{small}
  \end{center}
  \vskip -0.1in
\end{table*}

\subsection{Dataset}

To evaluate efficiency under images with high redundancy, we selected a high resolution medical dataset. FIVES \cite{FIVES} is a retinal blood vessel segmentation dataset consisting of $750$ fundus images, where each image is $2048{\times} 2048$ with full segmentation masks. While a goal of this work is segmentation performance under higher throughput, by dividing the fundus images into 4 quadrants, this increased the throughput by $10\times$ (${\sim} 10$ images per second to ${\sim} 100$  images per second at a 70\% keep ratio). Therefore, we operated on the quadrants rather than their original resolutions.  The dataset includes a diverse sample set of patients, including healthy patients and those diagnosed with Glaucoma, Diabetic Retinopathy and Age-related Macular Degeneration, in equal proportion. Additionally, approximately $200$ images are labelled as ``low quality'' with respect to colour distortion, blurring or contrast, so evaluation has a reasonable sample of mixed quality imaging. The divided images make up their own train/val/test splits with ratios (70:20:10).

To augment the images, we use standard transformations from the Albumentations library \cite{buslaev2020albumentations}. These include random horizontal and vertical flips (each with probability 0.5), random rotations by multiples of $90^{\circ}$ (probability 0.75), and colour jitter with brightness, contrast, and saturation factors up to 0.4 and hue up to 0.2. We further apply sharpening or unsharp masking (probability 0.4), Gaussian noise or Gaussian blur with kernel sizes between 3 and 7 (probability 0.4), and either random gamma adjustment in the range $[0.8, 1.2]$ or CLAHE \cite{zuiderveld1994contrast} (probability 0.3). Images are resized to the target resolution and normalised using ImageNet statistics \cite{deng2009imagenet}.


\subsection{Implementation Details}\label{sec:implementation}

For all experiments, we first pre-train an EoMT model \cite{kerssies2025your} initialised from a DINOv2 small checkpoint \cite{oquab2023dinov2}. Pre-training is performed for 50 epochs using tiered mask annealing over 4 epochs \cite{kerssies2025your}, starting at epoch 10. This model achieved 71.7\% AP on the test set. All subsequent experiments, including the non-pruned baseline results, are fine-tuned from this checkpoint to ensure a fair comparison.

Fine-tuning follows the same settings as pre-training, except that mask annealing is disabled and training is extended to 75 epochs. We use the AdamW optimiser \cite{loshchilov2017decoupled} with a learning rate of $2\times10^{-4}$ and weight decay of 0.05. Training is conducted with a batch size of 2, accumulated over 16 steps, while throughput data uses a batch size of 1. We apply an exponential moving average (EMA) with a decay of 0.9999, a layer-wise learning rate decay (LLRD) factor of 0.8, and use 16 query tokens applied over the final four transformer blocks. 
We adopt the same two-stage warmup followed by a polynomial decay learning rate schedule as used in EoMT \cite{kerssies2025your} using warmup steps of $[400, 800]$ for pre-training and $[200, 400]$ for fine-tuning, corresponding to non-backbone and DINOv2 backbone parameters respectively.

For routing and pruning, we set the routed token subset size $|\tau_r|$ to 50\% of the total tokens $N$, and sample the random routing range $[l\ldotsTwo n]$ uniformly with $l {\sim} \text{Unif} \{2\ldotsTwo \lfloor L/2\rfloor \}$ and $n {\sim} \text{Unif} \{l\ldotsTwo L{-}2 \}$. We chose 2 and $L{-}2$ from the fixed bound analysis from \cite{krause2025tread}, which also defines the bounds for our fixed routing results.  By default, pruning is applied from block $3$ to $L$ (zero-indexed), with the keep ratio value $\rho$ updated every three blocks. We set a base ratio of $\rho {=} 0.7$, resulting in the full schedule: $[ 0.7, 0.7, 0.7, 0.49, 0.49, 0.49, 0.34, 0.34, 0.34]$. Lastly, the policy loss weight $\lambda_{\text{pol}}$ is set to  8, as the standard Mask2Former loss is significantly larger in magnitude than $\mathcal{L}_{\text{pol}}$.

The model was trained on 4 NVIDIA H100s, which took several hours. All efficiency results were obtained on a single H100 using BF16 mixed precision tensors. We evaluate using standard metrics for binary segmentation and efficiency evaluation. These are the Dice coefficient, average precision and throughput. We do not evaluate against GFLOPs because it does not account for non-mathematical operational overheads such as gathers and scatters. For this reason, it is generally better to evaluate against real-world wall-clock time. Throughput is calculated by running the forward pass over 200 iterations on a randomly initialised tensor $I \in \mathbb{R}^{3\times 1024\times 1024}$, with the time starting after 20 warmup iterations.






\subsection{Evaluation \& Results}


We evaluate our proposed token routing strategy on binary segmentation under token pruning. Our baselines include a model trained without pruning, token merging over {\tt MHSA}, token merging over {\tt MHSA+MLP}, and a pruned baseline using the ratio loss $\mathcal{L}_{\text{ratio}}$. Token merging is applied over every block with a ratio set to a fixed 50\% of tokens by default as this is a similar throughput to the 70\% ratio of the pruned models. This can be seen directly in \cref{fig:AP/TH_TLDR}. These are compared against our pruned models with fixed bounds routing and random bounds routing {\paperacronym}. The results show that introducing routing consistently mitigates the depth-wise instability observed under pruning, leading to a substantial increase in average precision. Quantitative results are reported in \cref{tab:main_results}, while qualitative examples are shown in \cref{fig:qualitative_logit_routing_ablation}. We also explored variations in $\lambda_{\text{pol}}$ and $|\tau_r|$. While some settings gave slightly higher or lower average precision, the overall performance differences remained modest across the tested configurations. Lastly, we did not ablate the fixed token range $[l\ldotsTwo n]$, as \citet{krause2025tread} discover that longer routes than what we selected begin to hurt performance.


\subsubsection{Depth-wise Analysis}\label{sec:Depth_wise_Analysis}

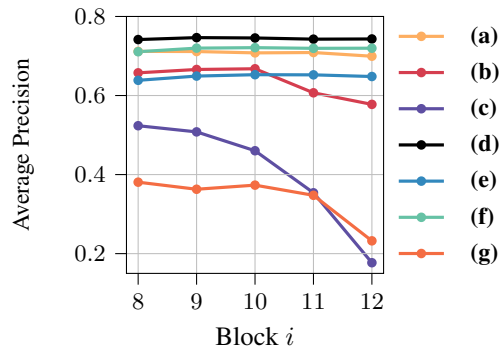
\begin{figure}
  \centering
    \def\BlockStart{12}
    
    \begin{tikzpicture}
    \begin{axis}[
        width=5cm,height=5cm,
        xlabel={Block $i$},
        ylabel={\small Average Precision},
        xtick pos=left,
        ytick pos=left,
        extra y ticks={0,1},
        xtick={8,9,10,11,12},
        xmin = 7.8,
        xmax = 12.2, 
        ymin = 0.15,
        ymax = 0.8, 
        solid,
        enlarge x limits=false,
        every x tick/.style={color=black, thin},
        every y tick/.style={color=black, thin},
        y tick label style={font=\small},
        x tick label style={font=\small},
        tick align=outside,
        xlabel near ticks,
        ylabel near ticks,
        axis on top,
        xmajorgrids,
        ymajorgrids,
        every axis plot/.append style={
            very thick,
            mark size=1.2pt
        },
        legend columns=1,
          legend style={
            at={(1.02,0.5)},      
            anchor=west,
            draw=none,
            font=\footnotesize,
            row sep=2pt,
            column sep=0.75em
          },
          legend cell align=left,
    ]
    

    \addplot[
          very thick,
          color=spectral4,
          mark=*
        ]
        table[
          col sep=comma,
          header=true,
          y index=4,
          x expr=13-\pgfplotstablerow
        ]{Graph_Data/Per_Block_PrunedMaskRouted.csv};
        \addlegendentry{\textbf{(a)}}\label{plots:Pruned_Mask_Routed_Depthwise}
    
        \addplot[
          very thick,
          color=spectral2,
          mark=*
        ]
        table[
          col sep=comma,
          header=true,
          y index=4,
          x expr=13-\pgfplotstablerow
        ]{Graph_Data/Per_Block_PrunedMaskFixedRouted.csv};
        \addlegendentry{\textbf{(b)}}\label{plots:Pruned_Mask_Fixed_Routed_Depthwise}

    \addplot[
      very thick,
      color=spectral11,
      mark=*
    ]
    table[
      col sep=comma,
      header=true,
      y index=4,
      x expr=13-\pgfplotstablerow
    ]{Graph_Data/Per_Block_PrunedMask.csv};
    \addlegendentry{\textbf{(c)}}\label{plots:Pruned_Mask_Depthwise}

    \addplot[
      very thick,
      color=black,
      mark=*
    ]
    table[
      col sep=comma,
      header=true,
      y index=4,
      x expr=13-\pgfplotstablerow
    ]{Graph_Data/Per_Block_NotPruned.csv};
    \addlegendentry{\textbf{(d)}}\label{plots:NotPruned_Depthwise}

    \addplot[
      very thick,
      color=spectral10,
      mark=*
    ]
    table[
      col sep=comma,
      header=true,
      y index=4,
      x expr=13-\pgfplotstablerow
    ]{Graph_Data/Per_Block_ToMe_MLP.csv};
    \addlegendentry{\textbf{(e)}}\label{plots:Attn_MLP_ToMe_Depthwise}

        \addplot[
      very thick,
      color=spectral9,
      mark=*
    ]
    table[
      col sep=comma,
      header=true,
      y index=4,
      x expr=13-\pgfplotstablerow
    ]{Graph_Data/Per_Block_ToMe.csv};
    \addlegendentry{\textbf{(f)}}\label{plots:Attn_ToMe_Depthwise}

    \addplot[
      very thick,
      color=spectral3,
      mark=*
    ]
    table[
      col sep=comma,
      header=true,
      y index=4,
      x expr=13-\pgfplotstablerow
    ]{Graph_Data/Per_Block_PrunedConst.csv};
    \addlegendentry{\textbf{(g)}}\label{plots:const_no_routing_Depthwise}


    \end{axis}
\end{tikzpicture}

\caption{
AP vs.\ convolutional decode block for each method. Results are from the same forward pass per checkpoint, with all efficient methods achieving ${\sim}100$ images/s throughput. Key: \textbf{(a)} {\paperacronym}, \textbf{(b)} Pruned (Fixed Routing), \textbf{(c)} Pruned (No Routing), \textbf{(d)} No Pruning, \textbf{(e)} ToMe \cite{bolya2023ToMeDiff} over {\tt MHSA+MLP}, \textbf{(f)} ToMe \cite{bolya2023ToMeDiff} over {\tt MHSA}, \textbf{(g)} Pruned with constant $\rho{=}0.5$ (No Routing).
}
\label{fig:depthwise}
\end{figure}

With optimisation strategies like token pruning, it is optimal to use the output from a block that achieves the best downstream performance at the best throughput. We could stop earlier in the model at specific block; however, it is unclear: which block should be selected a priori, and whether this will scale to deeper models. Besides, using multiple deconvolution steps introduces additional overhead.

To analyse depth-wise behaviour, we evaluate segmentation performance by plotting average precision across the 4 intermediate EoMT output blocks and after the final block \cref{fig:depthwise}. With default hierarchical keep-ratios and without routing, the performance consistently declines across these five blocks. The total active spatial tokens across those blocks are [2007, 2007, 1405, 1405, 1405] (out of a total of 4096), yet the observed performance decrease does not align with the reduction in token count. Therefore, we also compare against a constant ratio of 0.5 (\textbf{(g)} Fig. \ref{fig:depthwise}); but that generally performs worse compared to the hierarchical version by $\sim 12\%$ AP and shows the main performance decline at the final output. Moreover, when evaluating fixed bounds routing (\textbf{(c)} Fig. \ref{fig:depthwise}), the performance declines immediately after the fixed bounding ends, which motivates randomising the bounds. In addition, we do not observe the same depth-wise instability in the results of ToMe \cite{bolya2023ToMeDiff} (\textbf{(e,f)} Fig. \ref{fig:depthwise}), which is likely because it never uses tokens with stale representations. Although more experiments are needed to confirm, the decline in performance could be exacerbated by the use of LLRD. Additional depth-wise logit visualisations are provided in Appendix A (\cref{fig:qualitative_logit_depth-wise_ablation}).

\subsubsection{Throughput Analysis}
To analyse our method across a range of keep-ratios, we vary the keep ratio $\rho$ from 0.9 to 0.5. \cref{fig:AP/TH_TLDR} indicates this performance after the final block. To evaluate against other works or the non-pruned baseline, we plot against throughput rather than GFLOPs or the specific keep ratio. The figure indicates that without routing, the model experiences a sharp decline in performance between keep ratios 0.8 and 0.7. However, upon applying routing, the performance stabilises and increases.

\begin{figure}[t]
\centering
\begin{tikzpicture}

\pgfmathsetlengthmacro{\H}{0.25\columnwidth}             
\pgfmathsetlengthmacro{\KeyW}{0.5*(22/447)*\H + 0.2cm}  

\begin{groupplot}[
  group style={
    group size=4 by 1,
    horizontal sep=2.4mm,
    vertical sep=0pt
  },
  height=\H,
  width=\H,
  axis on top,
  scale only axis,
  tick align=outside,
  clip=false,
  tick label style={font=\scriptsize},
  xlabel style={font=\scriptsize},
  major tick length=1.5pt,
  xtick pos=bottom,
  ytick pos=left,
]

\nextgroupplot[
  xmin=0, xmax=13, ymin=0, ymax=13,
  y dir=reverse,
  trim axis left,
  xtick={0.5,2.5,4.5,...,12.5},
  ytick={0.5,2.5,4.5,...,12.5},
  xticklabels={0,2,4,...,12},
  yticklabels={0,2,4,...,12},
  xlabel={Layer Index},
]
\addplot graphics [xmin=0, xmax=13, ymin=0, ymax=13]
{Figs/Imgs/layer_similarity_test_routing=False_no_routing.png};

\nextgroupplot[
  xmin=0, xmax=13, ymin=0, ymax=13,
  y dir=reverse,
  trim axis left,
  xtick={0.5,2.5,4.5,...,12.5},
  ytick={0.5,2.5,4.5,...,12.5},
  xticklabels={0,2,4,...,12},
  yticklabels={},
  xlabel={Layer Index},
]
\addplot graphics [xmin=0, xmax=13, ymin=0, ymax=13]
{Figs/Imgs/layer_similarity_test_routing=True_fixed_routing.png};

\nextgroupplot[
  xmin=0, xmax=13, ymin=0, ymax=13,
  y dir=reverse,
  trim axis right,
  xtick={0.5,2.5,4.5,...,12.5},
  ytick={0.5,2.5,4.5,...,12.5},
  xticklabels={0,2,4,...,12},
  yticklabels={},
  xlabel={Layer Index},
]
\addplot graphics [xmin=0, xmax=13, ymin=0, ymax=13]
{Figs/Imgs/layer_similarity_test_routing=True_random_routing.png};

\nextgroupplot[
  width=\KeyW,
  xmin=0, xmax=1, ymin=0, ymax=1,
  xtick=\empty,
  ytick={0,0.2,0.4,0.6,0.8,1},
  yticklabels={0,0.2,0.4,0.6,0.8,1},
  ytick pos=right,
  axis lines=box,
  axis line style={black},
  tick label style={font=\tiny},
  major tick length=2pt,
  yticklabel style={xshift=2pt},
]
\addplot graphics [xmin=0, xmax=1, ymin=0, ymax=1]
{Figs/Imgs/layer_similarity_key.png};

\end{groupplot}
\end{tikzpicture}

\caption{Mean block-wise cosine similarity of tokens across the test set between EoMT layers under $\mathcal{L}_{\text{pol}}$ pruned models with the following routing configurations \textbf{(left)} no routing \textbf{(middle)} fixed bound routing \textbf{(right)} random routing.}\label{fig:cos_sim}
\end{figure}
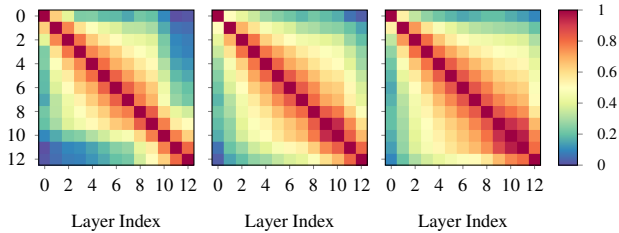



\subsection{Discussion}

\subsubsection{Routing Strategies}

We discussed the depth-wise results in \cref{sec:Depth_wise_Analysis}. Fixed routing bounds begin to mitigate the performance decay seen in the non-routing method. However, we still observe a decline in the model's performance after the maximum bound, showing that it overfits to the routing range. Alternatively, despite {\paperacronym} having the same maximum and minimum bounds as the fixed bounds model, it does not introduce the decline after that maximum bound. This suggests that random routing regularises tokens with respect to their active depth by exposing the model to many different routes. As seen in \cref{fig:cos_sim}, this encourages more consistent representations across blocks, therefore, leaving scope for fully skipping intermediate blocks in future work. 



\subsubsection{Informed Policies}

\begin{table}[t]
  \caption{Segmentation performance following the final block output for three choices of informed policy $\mathbf{T}$.}
  \label{tab:informed_policy_abl}
  \begin{center}
    \begin{small}
      \begin{sc}
        \begin{tabular}{lcc}
          \toprule
          Informed Policy $\mathbf{T}$             & Dice   & \begin{tabular}[c]{@{}l@{}}Average\\ Precision\end{tabular} \\ \midrule
        $\mathcal{L}_{\text{ratio}}$ & 90.46 & 71.67                                                      \\
        Segmentation mask            & 89.79 & 69.94                                                      \\
        Dist transform over mask & 89.89 & 70.20                                                      \\
        edges of mask                & 89.60 & 70.21                                                      \\ \bottomrule
        \end{tabular}
      \end{sc}
    \end{small}
  \end{center}
  \vskip -0.1in
\end{table}

Informed policies offer a different trade-off to ratio-based policies. While they offer inference at a range of compute budgets, they do this at a cost to performance. As in \cref{tab:informed_policy_abl}, the three choices we selected as informed policy did not meaningfully affect the results. In our experiments, we noticed that the rasterisation by downsampling our ground truth objectives $\mathbf{T}$ causes the pruning policy to activate much fewer tokens than expected from $\mathbf{T}$. Thus, likely leading to their decrease in performance. Despite the smaller number of activations, as a top-$k$ is applied, the actual total token count is the same. Given this, a hybrid ratio policy at early layers and a policy approach for later layers could be more effective. 


\subsubsection{Limitations}

This work has several limitations; first, our experiments are conducted on a single high resolution medical dataset. Although the depth-wise instability and routing behaviour are clear in this setting, further evaluation is required to assess its generality. Further, we observe that Token Merging \cite{bolya2023ToMeDiff} over just the {\tt MHSA} layer generally performs better out of the box. This is why we also compare our work to ToMe over {\tt MHSA+MLP}, as this is a closer match to the concept of pruning through cutting the {\tt MLP} computation in addition to the {\tt MHSA} layer. Additionally, we did not explore a learnt routing combination, like Mixture-of-Depths \cite{raposo2024mixture}, which would combine routing with the learnt policy outputs.




\section{Conclusion}

In this work, we observe an instability for token pruning methods that causes dense prediction performance to degrade at deeper layers. We attribute this to a train-test distribution shift when reactivating pruned tokens during inference. To mitigate this effect, we introduced token routing with random bounds, {\paperacronym}, which approximates the effect of pruning during training. In the future, we plan to explore the efficacy of {\paperacronym} for other dense prediction tasks such as depth estimation and 3D camera tracking.
We suspect this approach may complement existing efficiency strategies such as token merging \cite{bolya2023ToMe} and may also improve their application to deeper transformers. Furthermore, although we do not evaluate this setting, the observed behaviour motivates future work exploring its potential as a general regularisation mechanism, including in combination with dropout or stochastic depth \cite{dropout, stochasticdepth}.



\bibliography{pruning}
\bibliographystyle{icml2026}

\newpage
\appendix
\onecolumn
\section{Appendix}

\begin{figure*}[h]
\vspace*{0.25\textheight}
      \centering
      \begin{tikzpicture}
    
            
            \node[inner sep=0] (image1) at (0,0) {\includegraphics[width=0.95\linewidth]{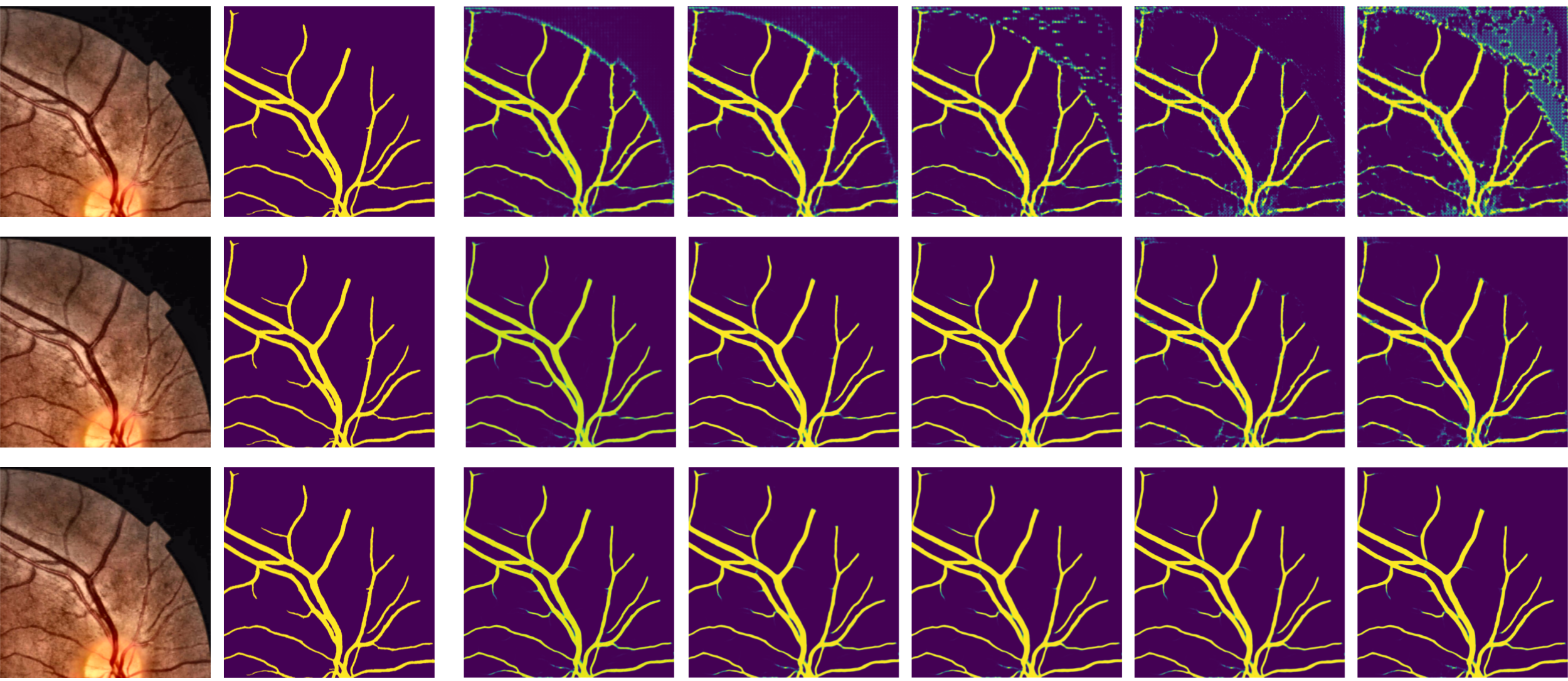}};

    
             \node[rotate=90, text width=2cm] at (-8.4,2.65)  {\footnotesize	 No Routing};
             \node[rotate=90, text width=2cm] at (-8.4,0.05)  {\footnotesize	 Fixed Routing};
             \node[rotate=90, text width=2cm] at (-8.4,-2.1)  {\footnotesize	 {\paperacronym}};

            \node[minimum height=0.5cm,text width=2cm] at (-6.5,3.75)  { Input};
            \node[minimum height=0.5cm,text width=2cm] at (-4,3.75)  {GT};
            
            \node[minimum height=0.5cm,text width=2cm] at (-1.8,3.75)  {Block 8};
            \node[minimum height=0.5cm,text width=3cm] at (1,3.75)  { Block 9};
            \node[minimum height=0.5cm,text width=3cm] at (3.25,3.75)  {Block 10};
            \node[minimum height=0.5cm,text width=3cm] at (5.6,3.75)  { Block 11};
            \node[minimum height=0.5cm,text width=3cm] at (7.9,3.75)  { Block 12};

        \end{tikzpicture}

      \caption{\textbf{Block-wise Qualitative segmentation results} - Block-wise segmentation logits after combining the weighted combination of logits across query masks. The top row shows the outputs after training with no routing. The middle row shows the fixed routing outputs, and the final row shows the model trained with {\paperacronym}.Each row contains a distinct input from the FIVES dataset \cite{FIVES}. Notably, the instability in row 1 is mitigated in rows 2 and 3. Row 3 also is able to produce less noisy outputs, especially around the optic disc. Further, the false-positive logits around the retinal field, introduced by pruning, are removed in both the fixed and random routing cases.}
      \label{fig:qualitative_logit_depth-wise_ablation}
\end{figure*}






\end{document}